\documentclass[runningheads]{llncs}
\usepackage[T1]{fontenc}
\usepackage{graphicx}
\usepackage{amsmath,amssymb,amsfonts,bm} 
\usepackage{booktabs,multirow,array} 
\usepackage[misc]{ifsym} 
\usepackage{hyperref}
\usepackage{color}

\usepackage{bm}
\usepackage{xcolor}
\newcommand{\legendbox}[1]{
\textcolor{#1}{\rule{1.2em}{1ex}}
}

\usepackage{multirow}
\usepackage{booktabs}
\usepackage{graphicx}

\begin{document}

\title{Single-Stage Hierarchical Rectification for \\
Weakly Supervised Histopathology Segmentation}
\titlerunning{SSHR for Weakly Supervised Histopathology Segmentation}

\author{
Duc T. Nguyen\inst{1,2,3} \and
Hoang-Long Nguyen\inst{1,2,3} \and
Thanh-Ha DO\inst{4} \and
Huy-Hieu Pham\inst{1,2,3}\thanks{Corresponding author: Huy-Hieu Pham (hieu.ph@vinuni.edu.vn).}
}

\authorrunning{D. T. Nguyen et al.}

\institute{
College of Engineering \& Computer Science, VinUniversity, Hanoi, Vietnam\\
\and
VinUni-Illinois Smart Health Center, VinUniversity, Hanoi, Vietnam\\
\and
The Computer Vision and Medical AI Lab, VinUniversity, Hanoi, Vietnam\\
\and
Posts and Telecommunications Institute of Technology, Hanoi, Vietnam
}

\maketitle              
\begin{abstract}
Existing weakly supervised semantic segmentation (WSSS)
methods in computational pathology rely on a multi-stage paradigm: class activation map (CAM) generation, offline pseudo-mask refinement, and fully supervised retraining. While established, this decoupled approach presents fundamental limitations. The multi-stage process not only incurs high computational training costs but also suffers from error propagation: local texture biases in shallow CNN layers generate false-positive artifacts that subsequent refinement steps often fail to correct. To address these persistent challenges through a simple yet highly effective approach, we propose the Single-Stage Hierarchical Rectification (SSHR) framework. Rather than passively refining CAMs post-hoc, our method proactively purifies intermediate feature representations during the forward pass. We introduce a Hierarchical Feature Rectification Module (HFRM) that utilizes deep global semantic context to filter out local anomalies in shallow layers. This mechanism generates high-fidelity activation maps directly within a single training loop. Experiments on the LUAD-HistoSeg and BCSS datasets demonstrate that SSHR outperforms state-of-the-art multi-stage methods. Furthermore, SSHR reduces training duration by 2 to 5 times. This efficiency minimizes computational overhead and accelerates clinical translation for large-scale histopathology workflows. The code is available at: \url{https://github.com/trongduc-nguyen/SSHR}.

\end{abstract}

\section{Introduction}
Deep learning has revolutionized computational pathology, offering automated solutions for tumor diagnosis, prognosis, and treatment planning \cite{srinidhi2021deep}. However, the success of these fully supervised models heavily relies on massive datasets with precise pixel level annotations \cite{van2019strategies}. In digital pathology, acquiring such dense annotations is exceptionally capital intensive and time consuming, as it requires highly specialized domain knowledge from experienced pathologists \cite{qu2022towards,wang2019weakly}. Consequently, the scarcity of pixel wise ground truth has become a major bottleneck, hindering the scalability of deep learning applications in clinical workflows \cite{amgad2022nucls}. To address this annotation burden, Weakly Supervised Semantic Segmentation (WSSS) has emerged as a promising paradigm, aiming to achieve segmentation performance comparable to fully supervised methods using only image level labels which are readily available in clinical reports \cite{montezuma2023annotating,abdelsamea2022survey}.

Mainstream approaches in histopathological WSSS typically rely on a multi-stage pipeline based on Class Activation Maps (CAMs)~\cite{zhou2016learning}. Early approaches focused on mitigating the noisy and sparse nature of CAMs caused by the morphological heterogeneity of tissue. For instance, HAMIL~\cite{zhong2023hamil} used a two stage framework with high resolution maps to refine boundaries. More recently, techniques like PBIP~\cite{tang2025prototype}, WAWEHIS~\cite{feng2025wave}, and ESFAN~\cite{zhang2025edge} have improved CAM quality through edge semantic synergy and prototype learning. Despite these advancements, existing techniques largely stay within a decoupled multi-stage workflow: (1) training a classification model, (2) applying complex post processing algorithms to refine CAMs into pseudo masks offline, and (3) retraining a separate segmentation network. This disjointed process inherently suffers from error propagation \cite{chong2021erase}. Shallow CNN layers provide sharp boundaries but possess limited receptive fields, making them highly susceptible to local texture variations. Consequently, the network frequently misclassifies isolated outlier tissues within dominant regions. When these artifacts are encoded into initial CAMs, the subsequent retraining step amplifies boundary errors rather than correcting them, leading to architectural instability. Furthermore, sequential pipelines incur heavy costs hindering clinical deployment \cite{hanna2025future}.

To address these challenges, we propose the Single-Stage Hierarchical Rectification (SSHR) framework. Unlike conventional methods that correct errors using external modules after CAM generation, SSHR intervenes directly at the feature level. We introduce an internal rectification mechanism that leverages the reliable global semantics of deep layers to dynamically filter local anomalies in shallow layers. By resolving this structural bias during the forward pass, SSHR generates precise activation maps in a unified training loop. The main contributions of our work are as follows:
\begin{itemize}
    \item[\textbullet] To the best of our knowledge, this is the first single-stage framework 
specifically designed for histopathological WSSS that outperforms 
multi-stage retraining pipelines. Our findings highlight the potential of internal rectification, which addresses the heavy dependency on the architecture during retraining where traditional refinement often fails to consistently improve segmentation quality beyond the initial CAMs.
    
    \item[\textbullet] We propose the Hierarchical Feature Rectification Module (HFRM), a dual branch mechanism combining global semantic filtering with spatial homogenization. By aligning low level details with semantic certainty, HFRM purifies CAMs directly during the forward pass.

    \item[\textbullet] SSHR outperforms existing state-of-the-art multi-stage weakly supervised segmentation methods on both the LUAD-HistoSeg and BCSS datasets. By eliminating offline mask generation and repeated retraining rounds, SSHR reduces total training duration by 2 to 5 times compared to conventional pipelines, facilitating rapid large scale clinical translation.
\end{itemize}


\section{Methodology}
\label{sec:methodology}

\subsection{Overall Framework}
Conventional Weakly Supervised Semantic Segmentation (WSSS) pipelines separate Class Activation Map (CAM) generation and boundary refinement, relying on delayed post-processing that often propagates false positives caused by local texture ambiguities in shallow CNN layers. To address this limitation, we propose the Single-Stage Hierarchical Rectification (SSHR) framework, which refines intermediate representations during the forward pass instead of correcting errors after CAM generation.

\begin{figure}[t]
    \centering
    \includegraphics[width=1\linewidth]{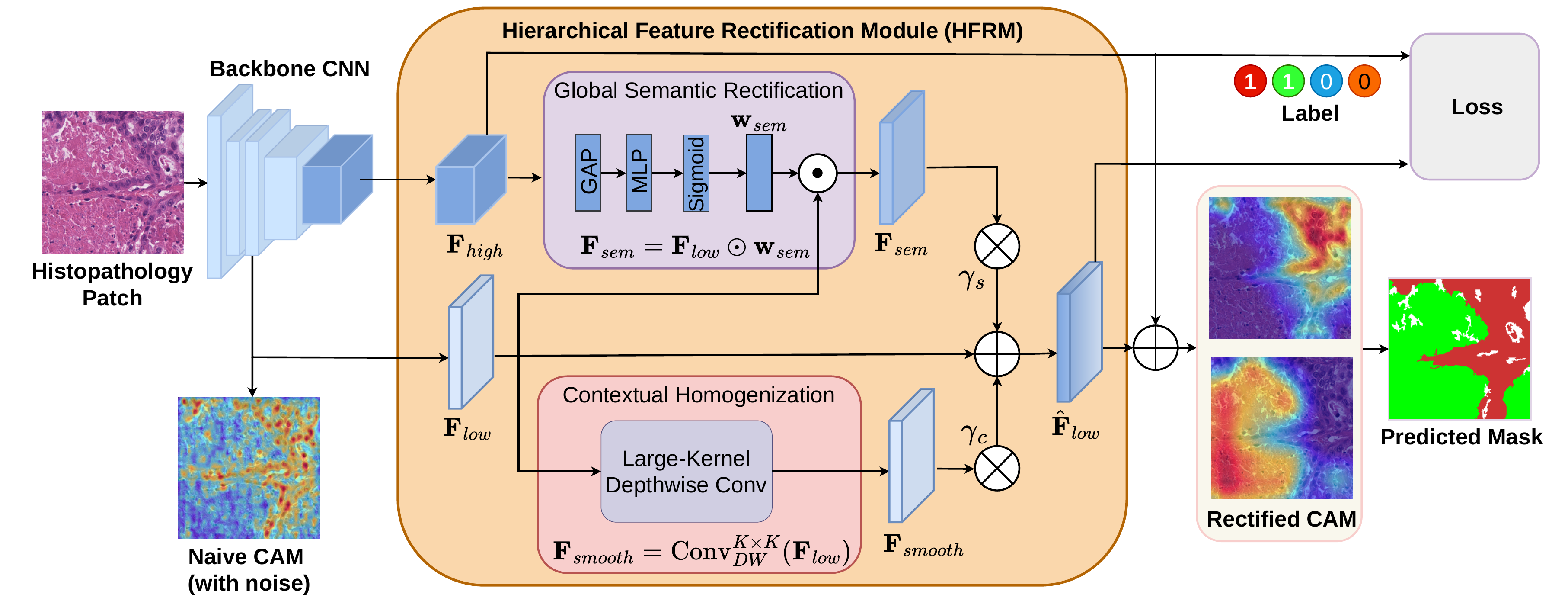}
    \caption{Overview of the proposed Single-Stage Hierarchical Rectification framework.}
    \label{fig:overview}
\end{figure}

As shown in Fig.~\ref{fig:overview}, the core component is the Hierarchical Feature Rectification Module (HFRM), which integrates semantic guidance with spatial homogenization. By rectifying features before activation maps are formed, SSHR suppresses local outliers and produces accurate CAMs without post-processing. The HFRM consists of two complementary branches described below.

\subsection{Global Semantic Rectification}
The Global Semantic Rectification (GSR) branch aims to harmonize the trade-off between local detail and global understanding. In histopathology, tissue structures exhibit high variety. While low level features are adept at finding clear edges, they often lack context, making them prone to local noise. Conversely, high level features capture the big picture but lack the necessary precision. The GSR branch bridges this gap by enforcing global consistency on low level representations, ensuring that fine details are guided by broader semantic information.

Let $\mathbf{F}_{low} \in \mathbb{R}^{C \times H \times W}$ denote the feature map from a low-level layer, and $\mathbf{F}_{high} \in \mathbb{R}^{D \times h \times w}$ denote the feature map from the deepest layer. Since $\mathbf{F}_{high}$ encodes the holistic image-level context, it contains inherent information about which tissue classes are globally present. We leverage this high-level prior to attenuate anomalous channels in $\mathbf{F}_{low}$.

We first apply Global Average Pooling (GAP) on $\mathbf{F}_{high}$ to obtain a global context vector $\mathbf{v}_{g} \in \mathbb{R}^{D}$. This vector is mapped through a lightweight multi-layer perceptron (MLP) and a Sigmoid activation to generate a channel-wise attention vector $\mathbf{w}_{sem} \in [0, 1]^{C}$:
\begin{equation}
    \mathbf{w}_{sem} = \sigma \left( \mathbf{W}_2 \delta (\mathbf{W}_1 \mathbf{v}_{g}) \right)
\end{equation}
where $\mathbf{W}_1 \in \mathbb{R}^{\frac{D}{8} \times D}$ and $\mathbf{W}_2 \in \mathbb{R}^{C \times \frac{D}{8}}$ are learnable projection matrices, and $\delta$ denotes the ReLU activation. The semantically rectified feature is computed via a Hadamard product along the channel dimension:
\begin{equation}
    \mathbf{F}_{sem} = \mathbf{F}_{low} \odot \mathbf{w}_{sem}
\end{equation}
This mechanism preemptively silences feature channels that would otherwise trigger false-positive predictions for absent classes.

\subsection{Contextual Homogenization}
While the semantic branch suppresses impossible classes at the image-level, spatial outliers belonging to valid classes can still persist locally. To enforce spatial contiguity, the second branch performs contextual homogenization using a lightweight large-kernel depthwise convolution, offering a computationally efficient alternative to self-attention.

Given a spatial kernel $K \times K$, the spatially smoothed feature $\mathbf{F}_{smooth} \in \mathbb{R}^{C \times H \times W}$ is obtained by:
\begin{equation}
    \mathbf{F}_{smooth} = \text{Conv}_{DW}^{K \times K}(\mathbf{F}_{low})
\end{equation}
By forcing each pixel to aggregate features from a broad neighborhood, isolated outlier pixels are structurally assimilated by the dominant surrounding tissue class, thereby eliminating erratic predictions.

To integrate these rectifications into a pre-trained backbone without triggering gradient shock, we formulate the final rectified feature $\hat{\mathbf{F}}_{low}$ using a zero-initialized residual connection:
\begin{equation}
    \hat{\mathbf{F}}_{low} = \mathbf{F}_{low} + \gamma_{s} \mathbf{F}_{sem} + \gamma_{c} \mathbf{F}_{smooth}
\end{equation}
where $\gamma_{s}$ and $\gamma_{c}$ are learnable scaling parameters. This non-invasive design ensures the network behaves like a standard CNN initially and progressively learns to balance native features with hierarchical structural corrections.

We apply $1 \times 1$ convolutions to the rectified intermediate features and the final high-level feature to generate scale-specific CAMs, $\mathbf{M}_s \in \mathbb{R}^{K \times H_s \times W_s}$, for $S$ different scales. Global Average Pooling extracts logits $\mathbf{p}_s$ to optimize the framework via an averaged multi-label soft margin loss:
\begin{equation}
    \mathcal{L}_{cls} = \frac{1}{S} \sum_{s=1}^{S} \left( -\frac{1}{K} \sum_{k=1}^K \Big[ y_k \log(\sigma(p_{s,k})) + (1-y_k) \log(1-\sigma(p_{s,k})) \Big] \right)
\end{equation}
where $y_k \in \{0, 1\}$ is the image-level label. During inference, the final high-resolution activation map $\mathbf{M}_{final}$ is obtained via a weighted fusion:
\begin{equation}
    \mathbf{M}_{final} = \sum_{s=1}^{S} \alpha_s \cdot \text{Norm}(\text{Up}(\mathbf{M}_s))
\end{equation}
where $\sum \alpha_s = 1$. The final segmentation mask is derived by taking the pixel-wise argmax of $\mathbf{M}_{final}$.

\section{Experiments}
\label{sec:experiments}
\subsection{Datasets and Experimental Settings}
\noindent \textbf{LUAD-HistoSeg:} This dataset contains whole slide images from 54 lung adenocarcinoma patients. It includes four tissue types: tumor epithelium (TE), tumor-associated stroma (TAS), necrosis (NEC), and lymphocyte (LYM). The dataset has 16,678 patches for training, 300 for validation, and 307 for testing \cite{han2022multi}.

    \noindent  \textbf{BCSS:} Derived from breast cancer whole slide images of 151 patients, this dataset also features four categories: tumor (TUM), stroma (STR), lymphocyte infiltration (LYM), and necrosis (NEC). It provides 23,422 patches for training, 3,418 for validation, and 4,986 for testing \cite{amgad2019structured}.
\begin{table*}[!ht]
\centering
\caption{Comparison with state-of-the-art methods on LUAD-HistoSeg and BCSS datasets. Phase 1 refers to initial CAMs/masks, while Phase 2 refers to the final segmentation after retraining. The results are reported as mean $\pm$ std. $^*$ indicates statistical significance compared with other methods 
($p < 0.05$, two-tailed $t$-test). Best results are highlighted in bold. }
\label{tab:luad_bcss_results}

\renewcommand{\arraystretch}{1.25}
\setlength{\tabcolsep}{6pt}
\resizebox{\textwidth}{!}{
\begin{tabular}{l l | cc | cc | cc}
\hline
\multirow{2}{*}{Method} &
\multirow{2}{*}{Venue} &
\multicolumn{2}{c|}{Backbone} &
\multicolumn{2}{c|}{mIoU (\%)} &
\multicolumn{2}{c}{mDice (\%)} \\
\cline{3-8}
 & & Phase 1 & Phase 2 & Phase 1 & Phase 2 & Phase 1 & Phase 2 \\
\hline

\multicolumn{8}{l}{\textbf{LUAD-HistoSeg}} \\
\hline

\multicolumn{8}{l}{\textit{Multi-Stage Methods}} \\
MLPS \cite{han2022multi} & MIA'22 & ResNet38 & ResNet101 & $74.18 \pm 0.89$ & $75.27 \pm 0.13$ & $85.42 \pm 0.73$ & $85.82 \pm 0.08$  \\
ARML \cite{feng2024mining} & MICCAI'24 & ResNet38 & ResNet101 & $73.44 \pm 0.27$ & $76.55 \pm 0.07$ & $84.65 \pm 0.18$ & $86.67 \pm 0.04$  \\
ESFAN \cite{zhang2025edge} & MICCAI'25 & ResNet38 & ResNet200 & $75.76 \pm 0.33$ & $76.83 \pm 0.38$ & $86.17 \pm 0.22$ & $86.86 \pm 0.24$  \\
PBIP \cite{tang2025prototype} & CVPR'25 & MiT-B1 & ResNet200 & $74.46 \pm 0.51$ & $75.89 \pm 0.26$ & $85.62 \pm 0.22$ & $86.02 \pm 0.13$  \\
WAWEHIS \cite{feng2025wave} & TMI'25 & ResNet38 & ResNet200 & $72.62 \pm 0.31$ & $74.69 \pm 0.52$ & $84.10 \pm 0.21$ & $85.46 \pm 0.32$  \\

\hline
\multicolumn{8}{l}{\textit{Single-Stage Methods}} \\
DuPL \cite{wu2024dupl} & CVPR'24 & ViT-B & -- & 63.71 $\pm$ 0.11 & -- & 73.06 $\pm$ 0.21 & --  \\
\textbf{SSHR (Proposed)} & -- & ResNet38 & -- & \textbf{77.93 $\pm$ 0.24$^*$} & -- & \textbf{87.54 $\pm$ 0.19$^*$} & --  \\
\hline

\hline
\multicolumn{8}{l}{\textbf{BCSS}} \\
\hline

\multicolumn{8}{l}{\textit{Multi-Stage Methods}} \\
MLPS \cite{han2022multi} & MIA'22 & ResNet38 & ResNet101 & $64.11 \pm 0.34$ & $69.07 \pm 0.35$ & $77.88 \pm 0.26$ & $81.47 \pm 0.28$  \\
ARML \cite{feng2024mining} & MICCAI'24 & ResNet38 & ResNet101 & $65.54 \pm 0.33$ & $69.58 \pm 0.51$ & $78.97 \pm 0.25$ & $81.83 \pm 0.37$  \\
ESFAN \cite{zhang2025edge} & MICCAI'25 & ResNet38 & ResNet200 & $70.34 \pm 0.40$ & $70.59 \pm 0.41$ & $82.41 \pm 0.23$ & $82.59 \pm 0.29$  \\
PBIP \cite{tang2025prototype} & CVPR'25 & MiT-B1 & ResNet200 & $66.74 \pm 0.34$ & $69.16 \pm 0.41$ & $79.06 \pm 0.31$ & $81.11 \pm 0.23$  \\
WAWEHIS \cite{feng2025wave} & TMI'25 & ResNet38 & ResNet200 & $64.63 \pm 0.54$ & $70.08 \pm 0.74$ & $78.27 \pm 0.42$ & $82.19 \pm 0.54$  \\

\hline
\multicolumn{8}{l}{\textit{Single-Stage Methods}} \\
DuPL \cite{wu2024dupl} & CVPR'24 & ViT-B & -- & 60.02 $\pm$ 0.51 & -- & 72.11 $\pm$ 0.49 & --  \\
\textbf{SSHR (Proposed)} & -- & ResNet38 & -- 
& \textbf{71.82 $\pm$ 0.38$^*$} 
& -- 
& \textbf{83.52 $\pm$ 0.27$^*$} 
& --  \\
\hline
\end{tabular}
}
\end{table*}
\subsection{Experimental Details}
 All experiments were conducted on a NVIDIA RTX 3090 GPU. We use a ResNet38 backbone  \cite{wu2019wider} for training. SGD optimizer with an initial learning rate of $1 \times 10^{-2}$ and a polynomial decay policy. We evaluate performance using two standard metrics: mean Intersection-over-Union (mIoU) and mean Dice coefficient (mDice).

\subsection{Comparison with state-of-the-art}

Table~\ref{tab:luad_bcss_results} compares SSHR with advanced WSSS methods. SSHR achieves the best performance on both datasets within a single training round. On LUAD-HistoSeg, it reaches 77.93\% mIoU, surpassing the final results of ESFAN (76.83\%) and PBIP (75.89\%), despite relying solely on a single-stage ResNet38 backbone, whereas these methods require heavier ResNet200 encoders during retraining. A similar advantage is observed on BCSS, where SSHR attains 71.82\% mIoU and consistently outperforms all multi-stage pipelines.

Evaluation of DuPL \cite{wu2024dupl}, a representative natural vision method, underscores pathology's severe domain shift. Despite its robust ViT-B architecture, DuPL suffers massive degradation as it lacks specialized mechanisms for the morphological heterogeneity and ambiguous transitions inherent in tissue. Unlike contour reliant natural image models, SSHR resolves structural biases via internal hierarchical rectification, effectively adapting the efficient single stage paradigm to pathological constraints.
\begin{figure}[!ht] 
    \centering
    \includegraphics[width=1\linewidth]{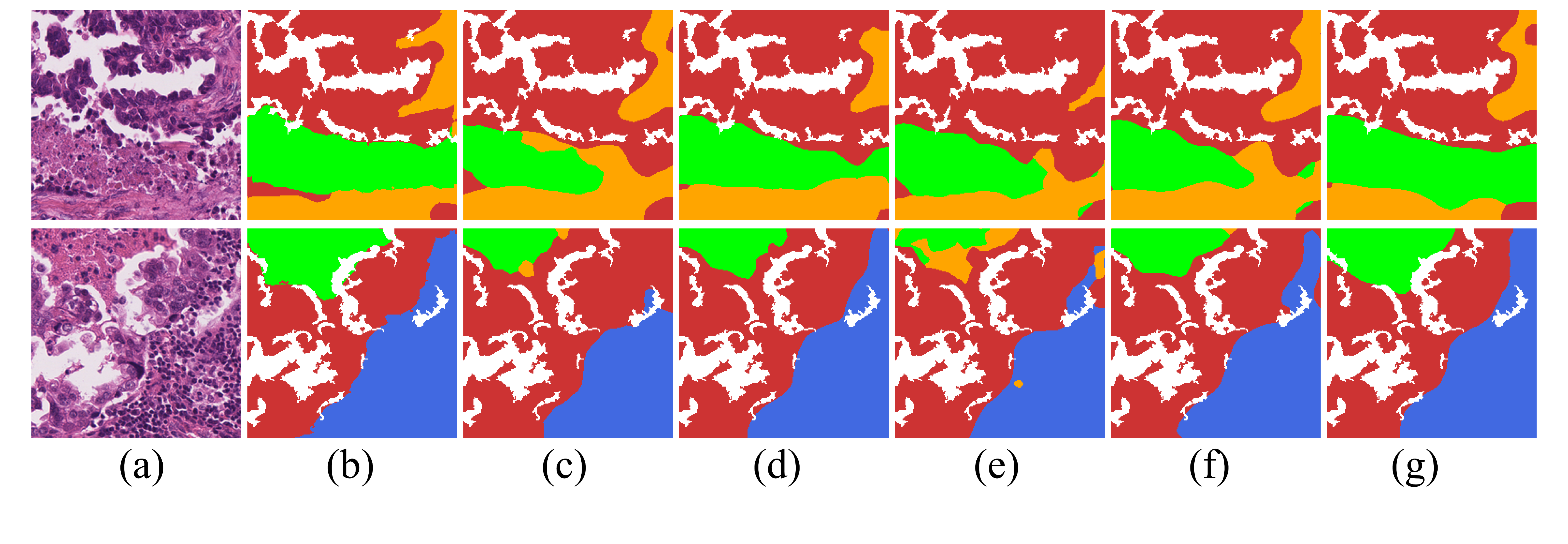}
    
    {\footnotesize
    LUAD-HistoSeg:
    \legendbox{red}~TE \;
    \legendbox{green}~NEC \;
    \legendbox{blue}~LYM \;
    \legendbox{yellow!70!orange}~TAS
    }
    
    \caption{Qualitative comparison of segmentation results. (a) Test image. (b) Ground truth. (c) ARML. (d) ESFAN. (e) MLPS. (f) WAWEHIS. (g) SSHR (Proposed method)}
    \label{fig:qualitative_results}
\end{figure}

As shown in Fig.~\ref{fig:qualitative_results}, SSHR achieves superior boundary adherence over multisstage methods. Internal hierarchical rectification effectively overcomes CAM coarseness, yielding masks precisely aligned with intricate tissue morphologies.

\subsection{Ablation Study}

\begin{table}[!ht]
    \centering
    \renewcommand{\arraystretch}{1.25}
    \caption{Ablation study evaluating the individual contributions of the proposed components. \textbf{BL}: Baseline.  The results are reported as mean $\pm$ std.}
    \label{tab:ablation_full}
    \begin{tabular}{c c c c | cc | cc}
        \toprule
        \multirow{2}{*}{\textbf{BL}} & 
        \multirow{2}{*}{\textbf{GSR}} & 
        \multirow{2}{*}{\textbf{CH}} & 
        \multirow{2}{*}{\textbf{$K$}} & 
        \multicolumn{2}{c|}{\textbf{mIoU (\%)}} & 
        \multicolumn{2}{c}{\textbf{mDice (\%)}} \\
        \cmidrule{5-8}
         & & & & LUAD-HistoSeg & BCSS & LUAD-HistoSeg & BCSS \\
        \midrule
        \checkmark & & & -- 
        & 72.12 $\pm$ 0.31 & 63.54 $\pm$ 0.41 
        & 82.83 $\pm$ 0.28 & 72.02 $\pm$ 0.37 \\

        \checkmark & \checkmark & & -- 
        & 75.84 $\pm$ 0.28 & 68.31 $\pm$ 0.35 
        & 84.41 $\pm$ 0.26 & 78.65 $\pm$ 0.32 \\

        \checkmark & & \checkmark & 15 
        & 75.21 $\pm$ 0.25 & 66.95 $\pm$ 0.39 
        & 84.88 $\pm$ 0.23 & 77.42 $\pm$ 0.36 \\
        \midrule
        \checkmark & \checkmark & \checkmark & 7 
        & 76.55 $\pm$ 0.20 & 70.42 $\pm$ 0.37 
        & 86.26 $\pm$ 0.18 & 79.08 $\pm$ 0.34 \\

        \checkmark & \checkmark & \checkmark & 15 
        & \textbf{77.93 $\pm$ 0.24} & \textbf{71.82 $\pm$ 0.38} 
        & \textbf{87.54 $\pm$ 0.19} & \textbf{83.52 $\pm$ 0.27} \\

        \checkmark & \checkmark & \checkmark & 21 
        & 75.41 $\pm$ 0.29 & 69.15 $\pm$ 0.49 
        & 84.02 $\pm$ 0.27 & 78.88 $\pm$ 0.44 \\
        \bottomrule
    \end{tabular}
\end{table}
We conduct an ablation study to evaluate the individual contributions of the proposed components (Table~\ref{tab:ablation_full}). Combining  Global Semantic Rectification (\textbf{GSR}) to suppress false positives and Contextual Homogenization (\textbf{CH}) to eliminate spatial outliers yields a strong synergistic performance boost over the baseline. Analyzing the spatial kernel size $K$ shows that $K=15$ optimally balances outlier suppression and boundary preservation. Smaller kernels ($K=7$) fail to eliminate large outliers due to limited receptive fields, while larger kernels ($K=21$) over smooth valid tissue boundaries, causing performance drops.

\subsection{Backbone Sensitivity and Instability in Stage 2}

\begin{table}[h]
\centering
\caption{Backbone scaling instability on LUAD-HistoSeg (mIoU \%). Red color and ${\color{red}\downarrow}$ denote cases where the performance in Phase 2 drops below Phase 1. The results are reported as mean $\pm$ std.} 
\label{tab:backbone_instability}

\renewcommand{\arraystretch}{1.25}
\setlength{\tabcolsep}{6pt}

\resizebox{\textwidth}{!}{
\begin{tabular}{l c | cccc}
\hline
\multirow{2}{*}{\textbf{Method}} 
& \multirow{2}{*}{\textbf{Phase 1}} 
& \multicolumn{4}{c}{\textbf{Phase 2 Backbone (mIoU \%)}} \\
\cline{3-6}
&  & ResNet101 & MiT-B4 & ResNet152 & EfficientNet-B8 \\
\hline
\textit{Params (M)} &  & 42M & 60M & 64M & 84M \\
\hline
MLPS \cite{han2022multi} & 74.18 $\pm$ 0.89
& 75.27 $\pm$ 0.18 $\uparrow$ 
& {\color{red} 72.61 $\pm$ 0.82 $\downarrow$} 
& {\color{red} 73.56 $\pm$ 0.37 $\downarrow$} 
& {\color{red} 70.55 $\pm$ 0.70 $\downarrow$} \\

ARML \cite{feng2024mining} & 73.43 $\pm$ 0.27
& 76.58 $\pm$ 0.06 $\uparrow$ 
& 73.53 $\pm$ 0.79 $\uparrow$ 
& 75.38 $\pm$ 0.25 $\uparrow$ 
& 74.71 $\pm$ 1.96 $\uparrow$ \\

ESFAN \cite{zhang2025edge} & 75.75 $\pm$ 0.33
& 75.76 $\pm$ 0.88 $\uparrow$ 
& {\color{red} 75.10 $\pm$ 0.62 $\downarrow$} 
& {\color{red} 75.61 $\pm$ 0.78 $\downarrow$} 
& {\color{red} 72.52 $\pm$ 0.28 $\downarrow$} \\

WAWEHIS \cite{feng2025wave} & 72.62 $\pm$ 0.31
& 75.41 $\pm$ 0.11 $\uparrow$ 
& {\color{red} 68.39 $\pm$ 1.26  $\downarrow$} 
& {\color{red} 66.91 $\pm$ 0.34 $\downarrow$} 
& 73.04 $\pm$ 0.70 $\uparrow$ \\

PBIP \cite{tang2025prototype} & 74.46 $\pm$ 0.51 
& {\color{red} 73.02 $\pm$ 0.17$ \downarrow$} 
& {\color{red} 72.94 $\pm$ 0.58$ \downarrow$} 
& 74.91 $\pm$ 0.21 $\uparrow$ 
& {\color{red} 73.18 $\pm$ 0.77 $\downarrow$} \\
\hline
\end{tabular}
}
\end{table}
Evaluating diverse Stage 2 backbones (Table \ref{tab:backbone_instability}) reveals inherent instability in sequential pipelines. Despite increased capacity, performance fluctuates unpredictably in multi stage frameworks, with methods like MLPS and PBIP often suffering mIoU degradation on heavier encoders or falling below Phase 1. These observations suggest that gains from retraining on noisy pseudo labels may depend on backbone characteristics and do not always yield systematic improvements, while Phase 1 errors can propagate during retraining. Conversely, SSHR eliminates this dependency by resolving local texture bias internally during the forward pass. By bypassing decoupled retraining, SSHR derives its accuracy from robust, rectified features, ensuring more stable performance compared to conventional 
multi-stage retraining pipelines.

\subsection{Computational Efficiency}
\begin{table}[h]
    \centering
    \renewcommand{\arraystretch}{1.25}
    \caption{Efficiency comparison on LUAD-HistoSeg. Ratio ($\times$) denotes total training time relative to Ours. Latency (ms) is the average inference time per patch.}
    \label{tab:efficiency}
    \begin{tabular}{lcccccc}
        \hline
        \textbf{Method} 
        & ESFAN \cite{zhang2025edge}
        & ARML \cite{feng2024mining}
        & WAWEHIS \cite{feng2025wave}
        & MLPS \cite{han2022multi}
        & PBIP \cite{tang2025prototype}
        & \textbf{Ours} \\
        \hline
        \textbf{Ratio ($\times$)} 
        & 2.23$\times$
        & 4.12$\times$
        & 3.85$\times$
        & 2.73$\times$
        & 5.42$\times$
        & \textbf{1.00$\times$} \\
        \textbf{Latency (ms)}
        & 20.25
        & 38.55
        & 24.45
        & 40.08
        & 22.75
        & \textbf{9.10} \\
        \hline
    \end{tabular}
\end{table}

Table~\ref{tab:efficiency} highlights the efficiency of SSHR, which reduces training time by $2.23\times$ to $5.42\times$ compared to multi stage methods. Due to the exceptionally large resolution of WSIs, which often reach gigapixel scales, training on large scale datasets in reality can potentially span weeks or even months. Such prolonged cycles in multi stage pipelines create severe bottlenecks for model development and clinical translation. By drastically cutting this duration, SSHR overcomes these temporal barriers, enabling rapid deployment in real world pathology workflows. Furthermore, SSHR achieves a superior inference latency of 9.10 ms per patch, outperforming all competitors and ensuring efficient inference.

\label{sec:conclusion}
\section{Conclusion}
SSHR highlights the potential of direct, on the fly CAM refinement as a robust and streamlined alternative to complex multi stage pipelines in pathology WSSS. By internalizing feature purification within a unified loop, SSHR avoids the error amplification typical of decoupled architectures. Our framework proves that leveraging high level semantics to rectify low level spatial features achieves superior accuracy while drastically reducing training duration. Without introducing additional inference overhead, SSHR provides an efficient solution for large scale clinical deployment in digital pathology.

\noindent \textbf{Limitations and Future Work:} SSHR depends on backbone depth for reliable semantics and currently lacks modeling of long range slide level interactions. Future efforts will focus on integrating global slide context and exploring spatially adaptive weighting strategies to further enhance robustness across staining variations and tissue morphologies in real world clinical environments.\\

\noindent \textbf{Acknowledgments.} This research was funded by the National Foundation for Science and Technology Development (NAFOSTED) through Project No. IZVSZ2\_229539 (2025–2027).\\

\noindent \textbf{Disclosure of Interests.} The authors have no competing interests to declare that
are relevant to the content of this article



\bibliographystyle{splncs04}
\bibliography{Paper-3170} 

@article{srinidhi2021deep,
  title={Deep neural network models for computational histopathology: A survey},
  author={Srinidhi, Chetan L and Ciga, Ozan and Martel, Anne L},
  journal={Medical image analysis},
  volume={67},
  pages={101813},
  year={2021},
  publisher={Elsevier}
}

@article{van2019strategies,
  title={Strategies to reduce the expert supervision required for deep learning-based segmentation of histopathological images},
  author={Van Eycke, Yves-R{\'e}mi and Foucart, Adrien and Decaestecker, Christine},
  journal={Frontiers in medicine},
  volume={6},
  pages={222},
  year={2019},
  publisher={Frontiers Media SA}
}

@article{qu2022towards,
  title={Towards label-efficient automatic diagnosis and analysis: a comprehensive survey of advanced deep learning-based weakly-supervised, semi-supervised and self-supervised techniques in histopathological image analysis},
  author={Qu, Linhao and Liu, Siyu and Liu, Xiaoyu and Wang, Manning and Song, Zhijian},
  journal={Physics in Medicine \& Biology},
  volume={67},
  number={20},
  pages={20TR01},
  year={2022},
  publisher={IOP Publishing}
}

@article{wang2019weakly,
  title={Weakly supervised deep learning for whole slide lung cancer image analysis},
  author={Wang, Xi and Chen, Hao and Gan, Caixia and Lin, Huangjing and Dou, Qi and Tsougenis, Efstratios and Huang, Qitao and Cai, Muyan and Heng, Pheng-Ann},
  journal={IEEE transactions on cybernetics},
  volume={50},
  number={9},
  pages={3950--3962},
  year={2019},
  publisher={IEEE}
}

@article{amgad2022nucls,
  title={NuCLS: A scalable crowdsourcing approach and dataset for nucleus classification and segmentation in breast cancer},
  author={Amgad, Mohamed and Atteya, Lamees A and Hussein, Hagar and Mohammed, Kareem Hosny and Hafiz, Ehab and Elsebaie, Maha AT and Alhusseiny, Ahmed M and AlMoslemany, Mohamed Atef and Elmatboly, Abdelmagid M and Pappalardo, Philip A and others},
  journal={GigaScience},
  volume={11},
  pages={giac037},
  year={2022},
  publisher={Oxford University Press}
}

@article{montezuma2023annotating,
  title={Annotating for artificial intelligence applications in digital pathology: A practical guide for pathologists and researchers},
  author={Montezuma, Diana and Oliveira, Sara P and Neto, Pedro C and Oliveira, Domingos and Monteiro, Ana and Cardoso, Jaime S and Macedo-Pinto, Isabel},
  journal={Modern Pathology},
  volume={36},
  number={4},
  pages={100086},
  year={2023},
  publisher={Elsevier}
}

@article{abdelsamea2022survey,
  title={A survey on artificial intelligence in histopathology image analysis},
  author={Abdelsamea, Mohammed M and Zidan, Usama and Senousy, Zakaria and Gaber, Mohamed Medhat and Rakha, Emad and Ilyas, Mohammad},
  journal={Wiley Interdisciplinary Reviews: Data Mining and Knowledge Discovery},
  volume={12},
  number={6},
  pages={e1474},
  year={2022},
  publisher={Wiley Online Library}
}

@inproceedings{zhou2016learning,
  title={Learning deep features for discriminative localization},
  author={Zhou, Bolei and Khosla, Aditya and Lapedriza, Agata and Oliva, Aude and Torralba, Antonio},
  booktitle={Proceedings of the IEEE conference on computer vision and pattern recognition},
  pages={2921--2929},
  year={2016}
}

@inproceedings{zhang2025edge,
  title={Edge-Semantic Synergy Fusion and Adaptive Noise-Aware for Weakly Supervised Pathological Tissue Segmentation},
  author={Zhang, Hualong and Feng, Siyang and Huan, Zihan and Wang, Huadeng and Liu, Zhenbing and Lan, Rushi and Pan, Xipeng},
  booktitle={International Conference on Medical Image Computing and Computer-Assisted Intervention},
  pages={160--169},
  year={2025},
  organization={Springer}
}

@inproceedings{feng2024mining,
  title={Mining gold from the sand: Weakly supervised histological tissue segmentation with activation relocalization and mutual learning},
  author={Feng, Siyang and Chen, Jiale and Liu, Zhenbing and Liu, Wentao and Wang, Zimin and Lan, Rushi and Pan, Xipeng},
  booktitle={International Conference on Medical Image Computing and Computer-Assisted Intervention},
  pages={414--423},
  year={2024},
  organization={Springer}
}

@article{feng2025wave,
  title={Wave-aware Weakly Supervised Histopathological Tissue Segmentation with Cross-scale Logits Distillation},
  author={Feng, Siyang and Zhang, Hualong and Zhao, Xianjing and Shi, Liting and Liu, Zhenbing and Lan, Rushi and Shi, Lei and Pan, Xipeng},
  journal={IEEE Transactions on Medical Imaging},
  year={2025},
  publisher={IEEE}
}

@inproceedings{tang2025prototype,
  title={Prototype-based image prompting for weakly supervised histopathological image segmentation},
  author={Tang, Qingchen and Fan, Lei and Pagnucco, Maurice and Song, Yang},
  booktitle={Proceedings of the Computer Vision and Pattern Recognition Conference},
  pages={30271--30280},
  year={2025}
}

@article{han2022multi,
  title={Multi-layer pseudo-supervision for histopathology tissue semantic segmentation using patch-level classification labels},
  author={Han, Chu and Lin, Jiatai and Mai, Jinhai and Wang, Yi and Zhang, Qingling and Zhao, Bingchao and Chen, Xin and Pan, Xipeng and Shi, Zhenwei and Xu, Zeyan and others},
  journal={Medical Image Analysis},
  volume={80},
  pages={102487},
  year={2022},
  publisher={Elsevier}
}

@article{zhong2023hamil,
  title={HAMIL: High-resolution activation maps and interleaved learning for weakly supervised segmentation of histopathological images},
  author={Zhong, Lanfeng and Wang, Guotai and Liao, Xin and Zhang, Shaoting},
  journal={IEEE Transactions on Medical Imaging},
  volume={42},
  number={10},
  pages={2912--2923},
  year={2023},
  publisher={IEEE}
}

@article{chong2021erase,
  title={Erase then grow: Generating correct class activation maps for weakly-supervised semantic segmentation},
  author={Chong, Yanwen and Chen, Xiaoshu and Tao, Yulong and Pan, Shaoming},
  journal={Neurocomputing},
  volume={453},
  pages={97--108},
  year={2021},
  publisher={Elsevier}
}

@article{hanna2025future,
  title={Future of artificial intelligence (AI)-machine learning (ML) trends in pathology and medicine},
  author={Hanna, Matthew G and Pantanowitz, Liron and Dash, Rajesh and Harrison, James H and Deebajah, Mustafa and Pantanowitz, Joshua and Rashidi, Hooman H},
  journal={Modern Pathology},
  pages={100705},
  year={2025},
  publisher={Elsevier}
}

@article{wu2019wider,
  title={Wider or deeper: Revisiting the resnet model for visual recognition},
  author={Wu, Zifeng and Shen, Chunhua and Van Den Hengel, Anton},
  journal={Pattern recognition},
  volume={90},
  pages={119--133},
  year={2019},
  publisher={Elsevier}
}

@article{amgad2019structured,
  title={Structured crowdsourcing enables convolutional segmentation of histology images},
  author={Amgad, Mohamed and Elfandy, Habiba and Hussein, Hagar and Atteya, Lamees A and Elsebaie, Mai AT and Abo Elnasr, Lamia S and Sakr, Rokia A and Salem, Hazem SE and Ismail, Ahmed F and Saad, Anas M and others},
  journal={Bioinformatics},
  volume={35},
  number={18},
  pages={3461--3467},
  year={2019},
  publisher={Oxford University Press}
}

@inproceedings{wu2024dupl,
  title={DuPL: Dual Student with Trustworthy Progressive Learning for Robust Weakly Supervised Semantic Segmentation},
  author={Wu, Yuanchen and Ye, Xichen and Yang, Kequan and Li, Jide and Li, Xiaoqiang},
  booktitle={Proceedings of the IEEE/CVF Conference on Computer Vision and Pattern Recognition},
  pages={3534--3543},
  year={2024}
}

\end{document}